\documentclass[journal, twoside]{IEEEtran}  

\IEEEoverridecommandlockouts                              

\usepackage{amsmath} 
\usepackage{amssymb} 
\usepackage{bbm} 
\usepackage{amsfonts}
\usepackage{mathtools}
\usepackage{times}  
\usepackage{balance}
\usepackage{bbold}
\usepackage[normalem]{ulem}

\usepackage{tabularx}

\usepackage{graphicx} 
\usepackage{blindtext}
\usepackage[hidelinks]{hyperref}
\usepackage{multirow}
\usepackage{tabularx}
\usepackage{booktabs}
\usepackage{subfig}
\usepackage{float}
\usepackage{color}
\usepackage{cite}
\usepackage[flushleft]{threeparttable}
\usepackage[linesnumbered, ruled]{algorithm2e}

\usepackage{geometry}
\def\equationautorefname~#1\null{(#1)\null}

\definecolor{axesRed}{RGB}{214,39,40}
\definecolor{axesGreen}{RGB}{44,160,44}
\definecolor{axesBlue}{RGB}{31,119,180}
\definecolor{myGold}{RGB}{255,192,0}

\title{
GMMLoc: Structure Consistent Visual Localization with Gaussian Mixture Models
}
\author{Huaiyang Huang$^1$, \textit{Student Member, IEEE}, Haoyang Ye$^1$, \textit{Student Member, IEEE}, \\ Yuxiang Sun$^1$, \textit{Member, IEEE}, and Ming Liu$^1$, \textit{Senior Member, IEEE} \\
\thanks{
Manuscript received February 24, 2020; revised May 17, 2020; accepted June 15,2020. This paper was recommended for publication by Editor Youngjin Choi upon evaluation of the Associate Editor and the Reviewers' comments.
This work was supported in part by the National Natural Science Foundation of China under Project U1713211, and in part by the Research Grant Council of Hong Kong under Project 11210017. \textit{(Corresponding author: Ming Liu.)}
}
\thanks{$^{1}$The authors are with \href{https://ram-lab.com/}{RAM-LAB}, the Hong Kong University of Science and Technology, Kowloon, Hong Kong. \textit{(Corresponding author: Ming Liu.)}
\texttt{\{hhuangat, hy.ye\}@connect.ust.hk}, \texttt{\{eeyxsun, eelium\}@ust.hk}.}
\thanks{Digital Object Identifier (DOI): 10.1109/LRA.2020.3005130.}
}

\begin{document}

\newcommand\Tstrut{\rule{0pt}{2.6ex}}         
\newcommand\Bstrut{\rule[-0.9ex]{0pt}{0pt}}   
\newcommand\todo[1]{\textcolor{yellow}{TODO: #1.}}         

\markboth{IEEE ROBOTICS AND AUTOMATION LETTERS. PREPRINT VERSION. ACCEPTED June, 2020} {Huang \MakeLowercase{\textit{et al.}}: Visual Localization with Gaussian Mixture Models}

\IEEEaftertitletext{\vspace{-1.0\baselineskip}}
\maketitle


\begin{abstract}
	Incorporating prior structure information into the visual state estimation could generally improve the localization performance.
	In this letter, we aim to address the paradox between accuracy and efficiency in coupling visual factors with structure constraints.
	To this end, we present a cross-modality method that tracks a camera in a prior map modelled by the Gaussian Mixture Model (GMM).
	With the pose estimated by the front-end initially, the local visual observations and map components are associated efficiently, and the visual structure from the triangulation is refined simultaneously.
	By introducing the hybrid structure factors into the joint optimization, the camera poses are bundle-adjusted with the local visual structure.
	By evaluating our complete system, namely GMMLoc, on the public dataset, we show how our system can provide a centimeter-level localization accuracy with only trivial computational overhead. In addition, the comparative studies with the state-of-the-art vision-dominant state estimators demonstrate the competitive performance of our method.
	\begin{center}
		\texttt{\url{https://github.com/hyhuang1995/gmmloc/}}
	\end{center}
\end{abstract}



\IEEEpeerreviewmaketitle

\section{INTRODUCTION}
\label{sec:intro}

\IEEEPARstart{L}{ocalization} is a crucial capability for robotic navigation,
as it can provide the global position and orientation which is essential for high-level applications ranging from path planning to decision-making \cite{liu2014topo}.
Among the available solutions for robot localization,
vision-based approaches are becoming increasingly popular due to
the widely-used low-cost and light-weight cameras \cite{desouza2002vision,mur2017orb}.
However, compared to ranging sensors, e.g., LiDARs, the shortcomings of the vision systems are not negligible in that, they generally measure the environment structure in an indirect way and suffer from large appearance variances of the environment \cite{pascoe2015direct}.

Integrating prior information from scene structures into visual localization systems could alleviate these issues.
Along this track, impressive results have been achieved in the recent work \cite{caselitz2016mloc, kim2018stereo, ding2018laser,huang2019metric,zuo2019visual,ye2020monocular}.
They usually adopt the pipeline that firstly builds a dense scene structure, and then localizes using visual or visual-inertial sensors with pre-built dense maps \cite{zuo2019visual}.
As the structure can be fully reused, this kind of \textit{modality-crossing} formulation between vision and structures allows the localization system to take advantage of both the rich features from visual sensors and the high-precision depth measurements from ranging sensors \cite{ye2020monocular}.
However, we observe that there is still a bottleneck on how to efficiently establish the constraints between structure-map elements and local visual measurements.
For example, building a kd-tree of a point cloud for searching the correspondences among triangulated visual landmarks takes logarithmic time with respect to the size of the data \cite{gold1998new}, while downsampling the map would cause a loss of information to a certain extent.

\begin{figure}[t!]
	\centering
	\includegraphics[width=0.48\textwidth]{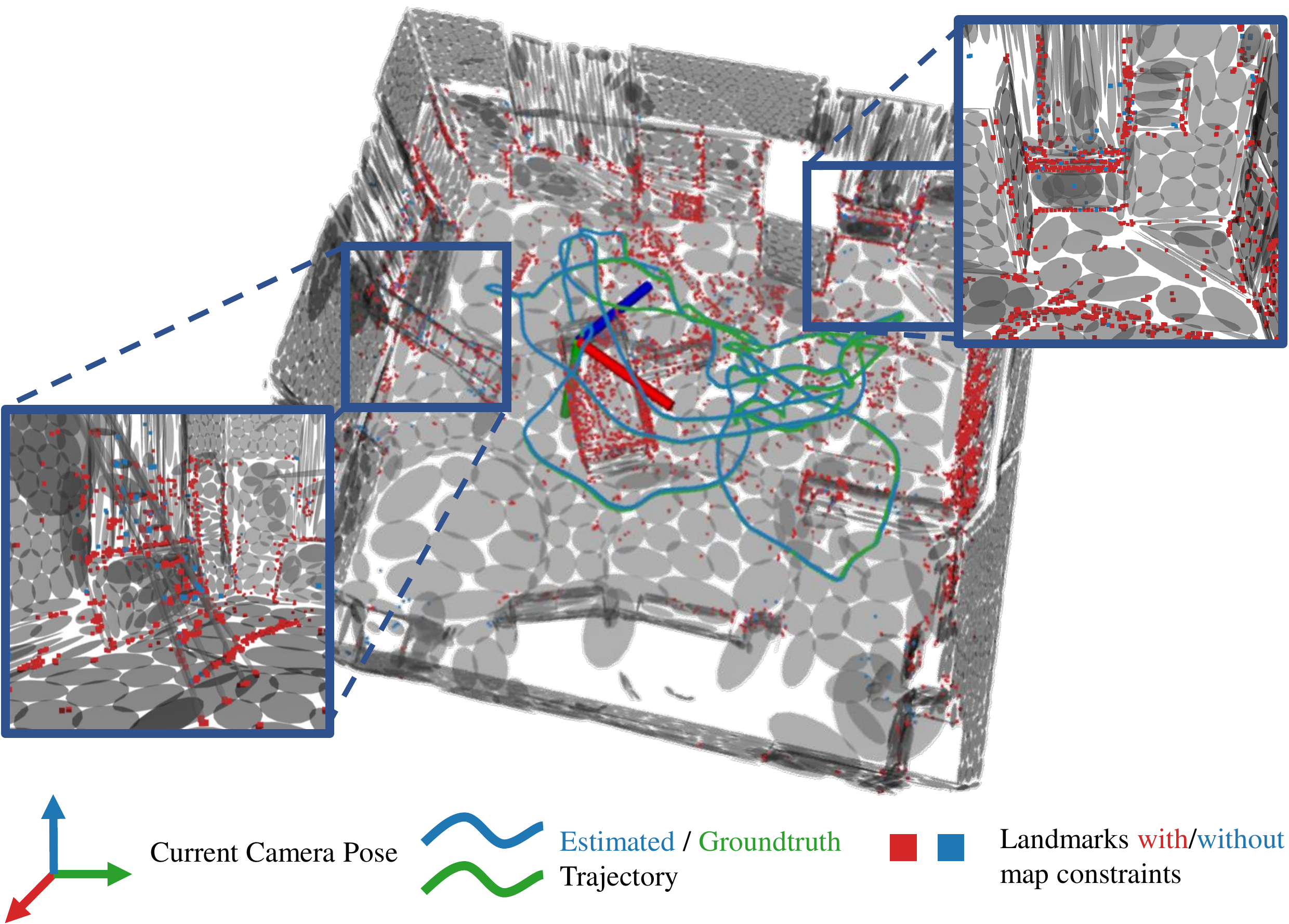}
	\caption{The proposed method localizes a camera (RGB-Axes) in a prior map represented by GMM. The GMM map is visualized as ellipsoids in $3\boldsymbol{\Sigma}$ bound. As shown in the global and zoom-in view, the estimated visual structure is well aligned with the global map and the trajectory is drift-less compared to the ground truth.}
	\vspace{-1em}
	\label{fig.exp}
\end{figure}

To resolve this issue, in this letter, we present a visual localization system modelling the prior distribution of the visual structure in 3-D space as a mixture of multivariate Gaussian distributions, namely the Gaussian Mixture Model (GMM) \cite{reynolds2009gaussian}.
As GMM is a much more compact data quantization method to represent a scene structure compared to the methods based on, for instance, the raw point cloud or voxel grid,
the structure information is naturally \textit{concentrated} into this parametric distribution with high-fidelity.
Initially tracking the camera pose by an indirect front-end, the proposed method associates the map components with local observations in a flash.
The structure constraints are established in a generic form,
with which landmarks from the triangulation are well refined.
Through optimizing camera poses and landmark positions in the joint Bundle Adjustment (BA), the structure factors are tightly coupled with visual factors from Multiview Geometry (MVG).
The experimental results show that the proposed approach achieves an accurate localization performance compared to the state-of-the-art methods, while only a trivial overhead is introduced.
A qualitative result is shown in \autoref{fig.exp}, where the local visual structure is well aligned with the GMM map, and the camera pose in the map frame is accurately recovered.
Demo videos and code can be found in our project homepage\footnote{\href{https://sites.google.com/view/gmmloc/}{https://sites.google.com/view/gmmloc/}} and github repository\footnote{\href{https://github.com/hyhuang1995/gmmloc}{https://github.com/hyhuang1995/gmmloc}}.
We summarize our contributions as follows:
\begin{enumerate}
	\item Representing the dense structure as GMM, we propose a hybrid structure constraint that ensures the global structure consistency in the visual state estimation.
	\item Following a hierarchical scheme, we further propose an efficient method that associates 3-D structure components with 2-D visual observations.
	\item Based on the proposed method, we implement GMMLoc, a novel visual localization system that tightly-couples the visual and structural constraints in a unified framework.
	\item Comparative experimental results demonstrate the remarkable performance of the proposed system. The additional study on reconstruction accuracy and structure factor supports our claims and confirms the effectiveness of the proposed method.
\end{enumerate}

\section{Related Works}
\label{sec:review}

\subsection{Visual Localization with Dense Prior Structure}
Visual localization is extensively pursued thus an exhaustive review is prohibitive. Here we limit our discussions to the methods which use the dense prior structure. Introducing dense prior structure has been shown to make a significant improvement to both robustness and accuracy in a vision-dominant localization system \cite{caselitz2016mloc,kim2018stereo,ding2018laser, huang2019metric, zuo2019visual, ye2020monocular}.
Caselitz \textit{et al}. \cite{caselitz2016mloc} proposed to associate the landmarks reconstructed from the monocular visual odometry \cite{mur2015orb} with a point-cloud map.
The 7-Degrees of Freedom (DoF) 
$\mathfrak{sim}(3)$ transform was estimated in an Iterative Closest Point (ICP) scheme.
Kim \textit{et al}. \cite{kim2018stereo} formulated the stereo localization as dense direct tracking of the disparity map with the local point-cloud.
Ding \textit{et al}. \cite{ding2018laser} proposed a sliding-window based stereo-inertial localization method with laser-map constraints. They introduced a hybrid optimization method to register the local sparse feature map with the prior laser map.
Huang \textit{et al}. \cite{huang2019metric} modelled the dense structure as an Euclidean Signed Distance Field (ESDF). The visual structure can then be aligned with the implicit surface.
Zuo \textit{et al}.\cite{zuo2019visual} proposed MSCKF with prior LiDAR map constraints (MSCKF w/ map). 
A Normal Distribution Transform (NDT)-based method was used to align the stereo reconstruction with the point cloud prebuilt from LiDAR.
Ye \textit{et al}. \cite{ye2020monocular} proposed DSL, where surfel constraints were introduced into the direct photometric error. The monocular camera can be localized in a tightly-coupled photometric BA framework.
While previous work succeeded in introducing structure constraints, either in a loosely-coupled or a tightly-coupled manner,
our method is different in that we quantize the geometry as a GMM, from which we formulate the structure constraint and introduce it into the visual state estimation. 
The advantages of GMM representation are two-fold:
first, it is highly compact, e.g., for the scene in \autoref{fig.exp}, the whole map consists of only 4500 Gaussian components, of which the data can be stored in an ASCII file of several kilobytes. Therefore it is efficient in both memory and storage;
second, such efficiency further accelerates the whole process for the association and establishing of the constraints.

\subsection{GMMs in State Estimation for Robotics}
Early probabilistic registration methods generally interpreted the point cloud data as GMMs by giving each point an isotropic Gaussian variance.
This paradigm was first proposed in \cite{gold1998new} to overcome the robustness issue of ICP \cite{besl1992method}
and its variants \cite{rusinkiewicz2001efficient}.
Later, Myronenko \textit{et al}. proposed the well-known CPD in \cite{myronenko2010point}, where a close-formed solution to the maximization step (M-step) of the EM algorithm was provided.
While these methods improved the robustness and accuracy, they are generally slower than ICP-based approaches.
To resolve this issue, recently, Gao \textit{et al}. proposed FilterReg \cite{gao2019filterreg}, where they formulated the expectation step (E-step) as a filtering problem and parameterized the point cloud data as permutohedral lattices.
Besides that, in \cite{eckart2018eccv}, Eckart \textit{et al}. provided an alternative solution by building a multi-scale GMM tree with anisotropic variances and 15-30 fps registration is achieved with the GPU. 
Similarly, aided by Inertial Measurement Unit (IMU) to recover roll and pitch, Dhawale \textit{et al}. \cite{dhawale2018fast} proposed a Monte Carlo localization method with the GMM based on the belief calculation given the depth map of a RGB-D camera.
They further showed that GMM can be an efficient environment modelling method for versatile navigation tasks, varying from occupancy analysis \cite{o2018variable} to exploration \cite{corah2019communication}.

Motivated by the success of these works, we assume the visual structure is subject to a probabilistic distribution over the Euclidean space, and formulate the constraints in a least-squares manner.
We further show how our method works in a vision-dominant localization system other than those based on ranging sensors \cite{rusinkiewicz2001efficient,myronenko2010point,gao2019filterreg, eckart2018eccv,dhawale2018fast, o2018variable,corah2019communication}.
Our method tightly-couples the structure factors with temporal visual factors, and 6-DoF motion parameters are fully recovered without IMU.

\begin{figure}[t!]
    \centering
    \includegraphics[width=0.45\textwidth]{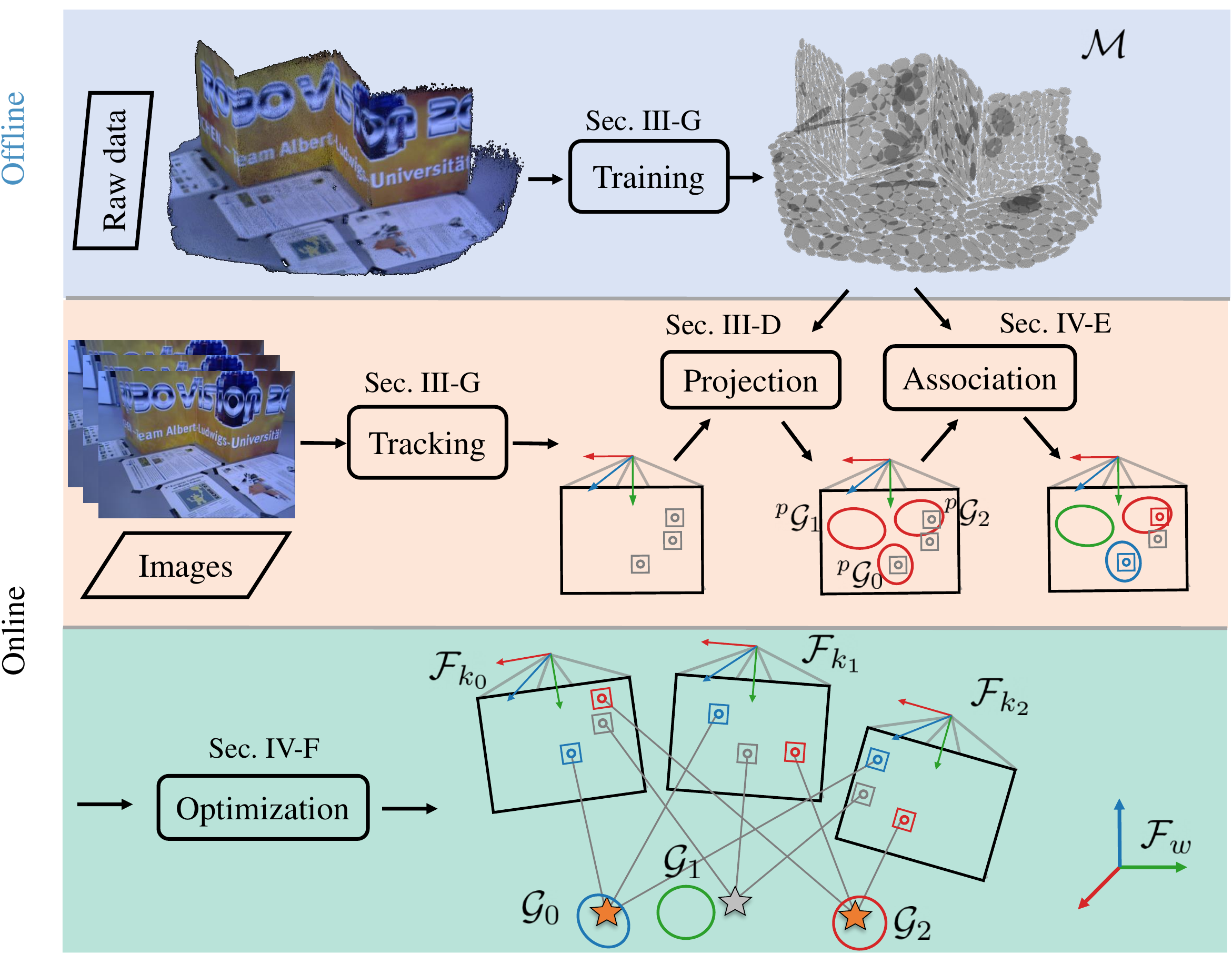}
    \caption{The flowchart of the proposed system. The model is processed offline. 
    After the visual tracking, the Gaussian components are projected to the image coordinate (shown as red ellipses).
    In the association step, candidates are first searched in 2-D. Then with the constraints from the map, the correspondence between the local measurement and map component is established (shown in the same color).
    The back-end optimization is a combination of hybrid constraints (shown in colors) and pure visual constraints (shown in grey).
    }
    \vspace{-1em}
    \label{fig.system}
\end{figure}
\section{Method}
The flowchart of our system is shown in \autoref{fig.system}, where we train the GMM offline from a given dense structure (e.g., point cloud).
Every input image is first tracked with a motion-only BA.
Then when a keyframe is selected, it will be utilized in the localization module.
For every keyframe, we project the GMM map back to the image coordinate.
Then local features are associated with map elements, and its corresponding landmark position from triangulation is refined simultaneously.
The joint BA optimizes the keyframe poses and the local structure, yielding a drift-less visual localization system.
The whole system introduces structural constraints and seamlessly maintains the global consistency.

\subsection{Notations}
Throughout the paper, the following notation is used: bold uppercase for the matrices , e.g., $\mathbf{R}$, bold lowercase for the vector e.g., $\mathbf{x}$, and light lower case for the scalar, e.g., $\theta$.

Given a point $\mathbf{x}\in \mathbb{R}^n$ and a Gaussian distribution $\mathcal{N}(\boldsymbol{\mu}, \boldsymbol{\Sigma})$, $\boldsymbol{\mu} \in \mathbb{R}^n$, $\boldsymbol{\Sigma} \in \mathbb{R}^{n\times n}$, Mahalanobis distance
$	\|\mathbf{x} - \boldsymbol{\mu}\|_{\boldsymbol{\Sigma}} = (\mathbf{x} - \boldsymbol{\mu})^T\boldsymbol{\Sigma}^{-1}(\mathbf{x} - \boldsymbol{\mu}) $
is used to measure the distance from the point to the distribution.
For measuring the distance between two Gaussian components $p(\mathbf{x} | \mathcal{G}_1) = \mathcal{N}(\boldsymbol{\mu}_1, \boldsymbol{\Sigma}_1)$, $p(\mathbf{x} | \mathcal{G}_2) = \mathcal{N}(\boldsymbol{\mu}_2, \boldsymbol{\Sigma}_2)$ in $\mathbb{R}^n$, we use the Bhattacharyya Coefficient (BC), which is given by:
\begin{equation*}
	dist_{n}(\mathcal{G}_1, \mathcal{G}_2) =\frac{1}{8}\|\boldsymbol{\mu}_{1}-\boldsymbol{\mu}_{2}\|_{\boldsymbol{\Sigma}}
	+\frac{1}{2} \ln \left(\frac{\operatorname{det} \boldsymbol{\Sigma}}{\sqrt{\operatorname{det} \boldsymbol{\Sigma}_{1} \operatorname{det} \boldsymbol{\Sigma}_{2}}}\right),
\end{equation*}
with $\boldsymbol{\Sigma} = (\boldsymbol{\Sigma}_{1}+\boldsymbol{\Sigma}_{2})/2$.

We denote the image collected at the $k$-th time as $I_k$ and the corresponding frame as $\mathcal{F}_k$.
For $\mathcal{F}_k$, the rigid transform $\mathbf{T}_k \in \mathbf{SE}(3)$ maps a 3-D point $\mathbf{x}_i \in \mathbb{R}^3$ in the world frame $\mathcal{F}_w$ to $\mathcal{F}_k$ using
$^c\mathbf{x}_k = \mathbf{R}_k \mathbf{p}_i + \mathbf{t}_k$,
where $\mathbf{T}_k = \left[ \mathbf{R}_k | \mathbf{t}_k \right]$. $\mathbf{R}_k$ and $\mathbf{t}_k$ are the rotational and translational components of $\mathbf{T}_k$, respectively.
Accordingly, $^c\mathbf{x}_k$ denotes a 3-D point in $\mathcal{F}_k$.
The camera pose, $\mathbf{T}_k$ is parameterized as $\boldsymbol{\xi}_k \in \mathfrak{se}(3)$.
We use $\pi : \mathbb{R}^3 \rightarrow \mathbb{R}^2$ to denote the projection function: $\mathbf{u} = \pi(^c\mathbf{x}_k) = \mathbf{K}{^c\mathbf{x}_k}$, where $\mathbf{u}$ is the projected pixel location in the image coordinate.
$\mathbf{K}$ stands for the intrinsic matrix.

\subsection{Problem Formulation}
The state, measurement and prior are defined as follows:
\subsubsection{state}
$\mathcal{X} = \mathcal{C} \cup \mathcal{L}$, where
$\mathcal{C} = \left\{\boldsymbol{\xi}_{0}, \boldsymbol{\xi}_{1}\dots \boldsymbol{\xi}_{m}\right\}$ is the set of keyframe poses in the local covisible map.
$\mathcal{L} = \left\{\mathbf{x}_{0}, \mathbf{x}_{1} \dots \mathbf{x}_{n}\right\}$ is the set of all the landmark positions.
While $\mathcal{X}$ stands for total states to be optimized,
we further denote the set of fixed keyframe poses as $\mathcal{C}^\prime = \left\{\boldsymbol{\xi}^\prime_{0}, \boldsymbol{\xi}^\prime_{1}\dots \boldsymbol{\xi}^\prime_{k}\right\}$, which serves as the prior information in the optimization.

\subsubsection{measurement}
The measurements consist of 2-D locations of landmarks observed in the pixel coordinate by the different keyframes, denoted as $\mathcal{Z} \doteq \left\{\mathbf{u}_{ik}\right\}_{(i, k)\in \mathcal{K}}$.
where $\mathbf{u}_{ik}$ is the pixel coordinate of the $i$-th landmark observed by $k$-th keyframe and $\mathcal{K}$ is the set of all the visual associations.
Similarly, we have $\mathcal{Z}^\prime \doteq \left\{\mathbf{u}_{ik}\right\}_{(i, k)\in \mathcal{K}^\prime}$, where $\mathcal{Z}^\prime$ represents the measurements associating prior keyframes with active landmarks.

\subsubsection{prior map}
The prior map is denoted as $\mathcal{M} = \left\{\mathcal{G}_0, \mathcal{G}_1, \dots \mathcal{G}_h\right\}$, where $\mathcal{G}_j \in \mathcal{M}$ stands for an individual map component (e.g., a voxel in a NDT-based method or a Gaussian distribution for GMMs).

The problem of visual localization against prior map can be formulated as a Maximum A Posteriori (MAP).
Instead of using pure visual or visual-inertial information, we introduce the constraints of the pre-built structure, which can be interpreted as defining a prior distribution of the observed visual structure, given as: $p(\mathcal{L}| \mathcal{M})$.
This leads the posterior to be factorized as follows:
\begin{equation*}
	p(\mathcal{X} | \mathcal{Z}, \mathcal{Z}^\prime, \mathcal{M}, \mathcal{C}^\prime) \propto  p(\mathcal{Z} | \mathcal{X}) \cdot p( \mathcal{L} | \mathcal{M}) \cdot
	p(\mathcal{Z}^\prime | \mathcal{C}^\prime, \mathcal{L})p(\mathcal{C})
\end{equation*}
\begin{equation*}
	=  \underbrace{\prod_{i,j} p(\mathbf{u}_{i, j} | \boldsymbol{\boldsymbol{\boldsymbol{\xi}}}_i, \mathbf{x}_j)}_{\text{visual factors}} \cdot
	\underbrace{\prod_{i, j} p(\mathbf{x}_{i} | \mathcal{G}_j)}_{\text{structure factors}} \cdot
	\underbrace{\prod_{i,j} p(\mathbf{u}^\prime_{i, j} | \boldsymbol{\xi}^\prime_i, \mathbf{x}_j)\prod_{i}p(\boldsymbol{\xi}_i)}_{\text{prior factors}}.
\end{equation*}

For the abundant advantages of GMMs as mentioned above,
here we model the prior structure as a generic GMM with anisotropic covariances:
\begin{equation}
	p(\mathcal{L} | \mathcal{M}) = \sum_{j = 0}^N w_j  p(\mathbf{x}_i|\mathcal{G}_j) = \sum_{j = 0}^N w_j  \mathcal{N}(\mathbf{x}_i; \boldsymbol{\mu}_j, \boldsymbol{\Sigma}_j).
\end{equation}
In other words, any landmark $\mathbf{x}_i$ should be subject to a prior distribution and its likelihood is given by
$ p(\mathbf{x}_i|\mathcal{G}_j) \propto \exp\left(\|\mathbf{x}_i - \boldsymbol{\mu}_j\|_{\boldsymbol{\Sigma}_j}\right)$.

Assuming the noise of measurements is zero-mean Gaussian, maximizing the posterior is equivalent to a least-squares optimization problem, with the objective function as follows:
\begin{equation}
	\label{eq.opt_total}
	E_\text{total} = E_\text{visual} + E_\text{structure} + E_\text{prior},
\end{equation}
where different residual terms are defined in the following section.

\subsection{Residual Definitions}
\label{sec.residual}

\subsubsection{Visual Factors}
Our system follows an indirect formulation of the visual residual, which is also known as the \textit{reprojection error}:
\begin{equation}
	\mathbf{e}_\text{proj}(\mathbf{x}_i, \boldsymbol{\xi}_k) = \mathbf{u}_{ik} - \pi(\mathbf{R}_{k}\mathbf{x}_i+\mathbf{t}_{k}),
\end{equation}
where $\mathbf{u}_{ik}$ is assumed with a Gaussian noise $\mathcal{N}\left(0, \boldsymbol{\Sigma}_{ik}\right)$, $\boldsymbol{\Sigma}_{ik} = \sigma_{ik}^2\mathbf{I}_{2\times2}$ and $\sigma_{ik}$ is the variance predefined for the local measurement.
Given the association set $\mathcal{K}$, visual factor $E_\text{visual}$ is given as:
\begin{equation}
	E_\text{visual} = \sum_{\left(i, k\right)\in \mathcal{K}} \rho\left(\|\mathbf{e}_\text{proj}\left(\mathbf{x}_i, \boldsymbol{\xi}_k\right)\|_{\boldsymbol{\Sigma}_{ik}} \right),
\end{equation}
where $\rho(\cdot)$ is the Huber norm for the robustness in the optimization.

\subsubsection{Structure Factors}

For a landmark $\mathbf{x}_i$ associated with a Gaussian component $\mathcal{G}_j$, the residual term can be derived from the Mahalanobis estimation.
Given the likelihood of $\mathbf{x}_i$ as
$ p(\mathbf{x}_i|\mathcal{G}_j)$,
maximizing the log-likelihood is equivalent to minimizing the Mahalanobis distance between $\mathbf{x}_i$ and $\mathcal{G}_j$, yielding the residual term:
\begin{equation}
	\mathbf{e}_\text{str} = \|\mathbf{x}_i - \boldsymbol{\mu}_j\|_{\boldsymbol{\Sigma}_j}.
\end{equation}

However, to make all the variables bundle-adjusted, calculating the likelihood or establishing constraints over all the components as probabilistic registration methods \cite{myronenko2010point} is not applicable.
Additionally, constraining the landmark position with a 3-D component can somehow be ``misleading", as it attempts to minimize the distance from the landmark to the mean.
Inspired by \cite{eckart2018eccv}, where the authors propose to decompose the anisotropic covariance for accelerating the Mahalanobis estimation, we introduce a hybrid objective function based on the degeneration of different components.
As a real-world scene structure is generally constructed by planars, we observe that in a GMM with anisotropic variance directly fitted from a dense point cloud, many components tend to degenerate.
Therefore, we detect the degeneration of 3-D Gaussian components in the preprocessing step.

Decomposing the covariance via SVD gives $\boldsymbol{\Sigma}_j = \mathbf{USV}^T$.
For $\mathbf{U} = \left[\mathbf{e}_1, \mathbf{e}_2, \mathbf{e}_3\right]$,
we further let $\mathbf{e}^\prime_3 = \mathbf{e}_1 \times \mathbf{e}_2$ just to ensure it meets the right-hand rule as commonly used in our system, and denote $\mathbf{R} = \left[\mathbf{e}_1, \mathbf{e}_2, \mathbf{e}_3^\prime\right]$.
Due to the orthonormality between different eigenbases $\mathbf{e}_i$, we have $\mathbf{R} \in \mathbf{SO}(3)$.
As the covariance matrix $\boldsymbol{\Sigma}_j$ is symmetric and positive definite,
we have $\mathbf{U}\equiv\mathbf{V}$.
Accordingly, the factorization of $\boldsymbol{\Sigma}_j$ is rewritten as:
\begin{equation}
	\boldsymbol{\Sigma}_j = \mathbf{RSR}^T,
	\mathbf{S} = \text{diag}(\lambda_1, \lambda_2, \lambda_3), \mathbf{R} = [\mathbf{e}_1, \mathbf{e}_2, \mathbf{e}^\prime_3],
\end{equation}
where $\mathbf{S} = \text{diag}(\lambda_1, \lambda_2, \lambda_3)$, $\lambda_1 < \lambda_2 < \lambda_3$ is the diagonal matrix of singular values.
In a geometric interpretation, $\mathbf{R}$ is equivalent to the rotation part of the transform from the component coordinate to the world coordinate.
In addition a singular value $\lambda_i$ also stands for the scaling factor according to the data distribution along $\mathbf{e}_i$.
A small $\lambda_1$ indicates $\boldsymbol{\Sigma}_j$ tends to be degenerated in rank,
or in a geometric interpretation, the component is more similar to a planar.
We use $\mathbb{1}(\mathcal{G}_j)$ to indicate whether the i-th component is degenerated or not, given by:
\begin{equation}
	\mathbb{1}(\mathcal{G}_j) =
	\begin{cases}
		1 & \lambda_1 \ll \lambda_2 < \lambda_3 \\
		0 & \text{otherwise}                    \\
	\end{cases}.
\end{equation}
For the degenerated case, we consider the residual term as follows:
\begin{equation}
	e_\text{str\_deg}(\mathbf{x}_i, \mathcal{G}_j) = \|\mathbf{e}_{j1}(\mathbf{x}_i - \boldsymbol{\mu}_j)\|_{\boldsymbol{\Sigma}_\text{str}},
\end{equation}
where $\boldsymbol{\Sigma}_\text{str} = \sigma^2_\text{str}\mathbf{I}_{3\times3}$, $\sigma_\text{str}$ is the pre-defined variance of structure constraints, which can also be interpreted as a coupling factor for balancing the visual and structural constraints. The effect of $\sigma_\text{str}$ is further discussed in \autoref{sec.exp_param}.
Intuitively, this formulation can be explained as point-to-plane distances, which is efficient for computation
and provides a more geometrically explainable formulation of the constraint. 
Denote the association set as $\mathcal{S}$, the total structure objective function is given by:
\begin{equation}
	\begin{split}
		E_\text{structure} = &\sum_{(i, j)\in \mathcal{S}} (\mathbb{1}(\mathcal{G}_j)e_\text{str\_deg}(\mathbf{x}_i, \mathcal{G}_j) \\
		&+ \left(1 - \mathbb{1}(\mathcal{G}_j)\right) e_\text{str}(\mathbf{x}_i, \mathcal{G}_j))
	\end{split}
\end{equation}


\subsubsection{Prior Factors}
In addition to fixed keyframe poses in some visual factors, here we discuss how we deal with the initial estimation.
As the proposed method does not aim to solve a global retrieval problem,
we consider a prior pose is given at the initialization of the system.
In detail, two conditions are discussed:
\begin{itemize}
	\item if an accurate pose is given (e.g., re-localization from feature map), we set it to the first frame and fix it in the optimization.
	\item if an initial guess is provided (e.g., from manually assigned), a prior term for constraining the initial pose is added to the optimization, defined as:
	      \begin{equation}
		      \mathbf{e}_\text{init} = \log(\exp(\boldsymbol{\xi}_\text{init}^\land)^{-1}\exp\left(\boldsymbol{\xi}^{\land}_{c_0}\right))^\lor,
	      \end{equation}
	      where $\boldsymbol{\xi}_\text{init}$ and $\boldsymbol{\xi}_{c_0}$ are the preset initial guess and the actual pose for the initial keyframe.
\end{itemize}


\begin{figure}[t!]
	\centering
	\subfloat[the GMM map.]{\includegraphics[width=0.23\textwidth]{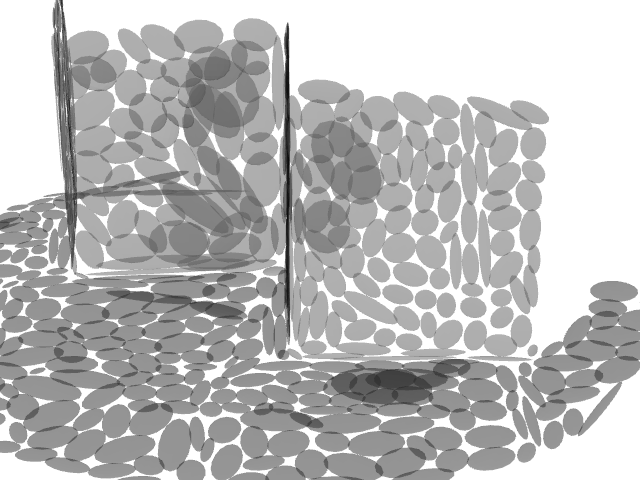}}
	\quad \subfloat[an image view.]{\includegraphics[width=0.23\textwidth]{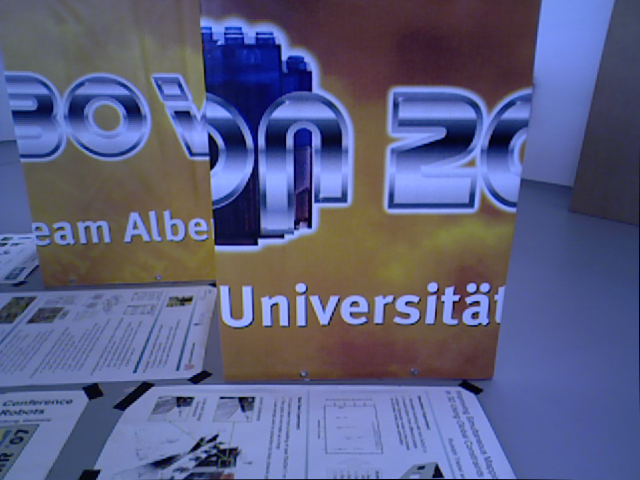}}\\
	\subfloat[direct projection result.]{\includegraphics[width=0.23\textwidth]{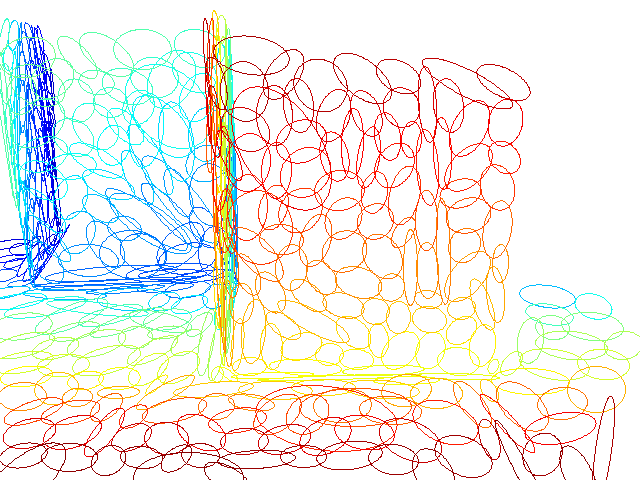}}
	\quad \subfloat[result with post-processing.]{\includegraphics[width=0.23\textwidth]{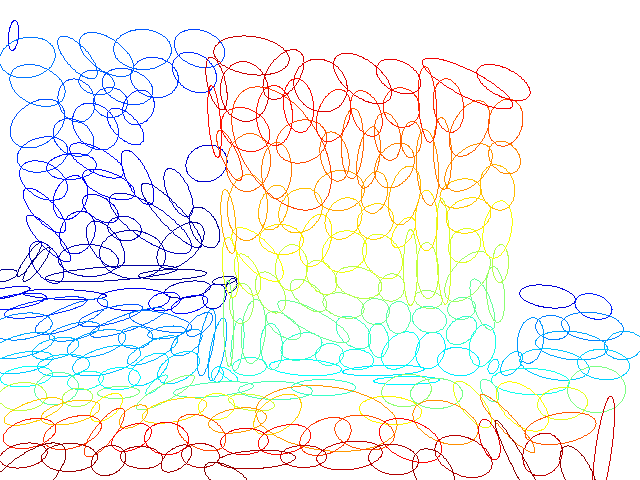}}
	\caption{An example of the GMM projection from the map to the image at the same viewpoint, colored by depth. The filtered result generally has fewer mis-projected components.}
	\vspace{-1em}
	\label{fig.projection}
\end{figure}

\subsection{Projection of the GMM Map}
\label{sec.proj}

When a keyframe is created, we assume its pose $\boldsymbol{\xi}_k$ is tracked and local observations $\mathcal{O}_k \doteq \left\{\mathbf{u}_{ik}\right\}_{i=1\dots n} $ are detected.
To associate the map elements with local observations, we first project the Gaussian components to the image coordinates.
With the camera pose $\boldsymbol{\xi}_k$ recovered by the tracking frontend, this projection process can be regarded as a nonlinear transformation of the Gaussian components.
Similar discussion can be found in \cite{barfoot2019state, dhawale2018fast}.
As the transformation of a point in 3-D Euclidean space is a linear operation, for an individual component $\mathcal{G}_j$, $p(\mathbf{x}_i | \mathcal{G}_j) = \mathcal{N}(\boldsymbol{\mu}_j, \boldsymbol{\Sigma}_j)$, the density function under $\mathcal{F}_k$ is:
\begin{equation}
	p(^{c_k}\mathbf{x}_i | ^c\mathcal{G}_j) = \mathcal{N}(\mathbf{R}_k\boldsymbol{\mu}_j + \mathbf{t}_k, \mathbf{R}_k\boldsymbol{\Sigma}_j\mathbf{R}_k^T),
\end{equation}
while the projection function $\pi(\cdot)$ is nonlinear due to the implicit normalization of the point under image coordinate.
A first-order approximation gives:
\begin{equation}
	p(\mathbf{u} | ^{p}\mathcal{G}_j) = \mathcal{N}\left(\pi\left(^{c}\boldsymbol{\mu}_j\right), \left.\mathbf{J}_\pi\right|_{^c\boldsymbol{\mu}_j} \mathbf{R}_k\boldsymbol{\Sigma}_j\mathbf{R}_k^T\mathbf{J}_\pi^T|_{^{c}\boldsymbol{\mu}_j}\right)
\end{equation}
where we denote a 2-D Gaussian component projected from $\mathcal{G}_j$ in the pixel coordinate as $^p\mathcal{G}_j$. ${^{c}}\boldsymbol{\mu}_j = \mathbf{R}_k\boldsymbol{\mu}_j + \mathbf{t}_k$ is the transformed mean vector and $\mathbf{J}_{\pi}$ is the Jacobian of $\pi(\cdot)$ with respect to ${^{c}\boldsymbol{\mu}_j}$. An example projection result is shown in \autoref{fig.projection}.

As we manually ``render" the scene with CPU, to generate a photorealistic projection result,
we generally use the following criteria to filter the projected components:
\begin{itemize}
	\item Check if $\mathcal{G}_j$ lies within the image frustum by $(^c\boldsymbol{\mu}_j)_z > 0$.
	\item For the degenerated component, the angle between viewing ray and $\mathbf{e}_{j1}$ (degenerated axis) of $\mathcal{G}_j$ is checked. If
	      \begin{equation*}
		      \frac{\langle\mathbf{t}_{k} - \boldsymbol{\mu}_{j}, \mathbf{e}_{j1}\rangle}{\|\mathbf{t}_{k} - \boldsymbol{\mu}_{j}\|} < \cos \delta_\theta,
	      \end{equation*}
	      the component is not observable by the current frame.
	\item The $\boldsymbol{\Sigma}_{2\times2}$ of $^p\mathcal{G}_j$ is decomposed, and if its singular value $\lambda_{j1} < \lambda_{j2} \ll \delta_\lambda$, the component is considered less representative and is therefore discarded.
	\item For the remainder we check the occlusion condition.
	      For $ \hat{^p\mathcal{G}_j} = {\arg \min}_{^p\mathcal{G}_j} dist_2(^p\mathcal{G}_i, ^p\mathcal{G}_j)$, if $(^c\boldsymbol{\mu}_i)_{z} < (^c\boldsymbol{\mu}_j)_z$, $\mathcal{G}_j$ is considered a background component and is supposed to be occluded by the foreground component.
\end{itemize}
An example filtered result is shown in \autoref{fig.projection}.
The whole projection procedure for a common scene, e.g., the one shown in \autoref{fig.exp}, can be efficiently finished in several milliseconds using CPU only.

\subsection{Structure Optimization and Association}
The complete method for associate local observations with map elements is shown in \autoref{alg.alg_association}.
Given a keypoint $\mathbf{u}_{ik}$ that can be successfully triangulated either from temporal or static stereo,
we select $k$-nearest 2-D Gaussian components as association candidates, from the set of current projection results $\mathcal{P}$, where the distance metric defined by $\|\mathbf{u}_{ik} - \boldsymbol{\mu}_j \|_{\boldsymbol{\Sigma}_j}$.
This gives the candidate set $\mathcal{P}_i\doteq\left\{^p\mathcal{G}_j\right\}, |\mathcal{P}_i| = k$ (line 1).
We then optimize the newly generated landmark position and find the best-fit component (line 2-9).
With the triangulated position $\mathbf{x}_i$, we iterate over $\mathcal{P}_i$,
and define a sub-problem for optimizing $\mathbf{x}_i$:
\begin{equation}
	\label{eq.str_opt}
	\hat{\mathbf{x}}_i= \arg \min_{\mathbf{x}_i} \left(\sum_{k^\prime} \|\mathbf{e}_\text{proj}\left(\mathbf{x}_i, \boldsymbol{\xi}_{k^\prime}\right)\|_{\boldsymbol{\Sigma}_{ik^\prime}} + e_\text{str}\left(\mathbf{x}_i, \mathcal{G}_j\right)\right).
\end{equation}
After the optimization, visual residuals are checked to discard outliers.
The threshold for visual factors $th$ is determined by $\mathcal{X}^2$-test, where if the $e_\text{proj} > th$, we consider this association invalid (line 4-8).
Assuming that a valid association is found,
we denote the optimal position with the constraint from $^p\mathcal{G}_j$ as $\hat{\mathbf{x}}_i^j$.
We select the final association $\hat{^p\mathcal{G}_j}$ leading minimum reprojection error (line 7):
$$\hat{^p\mathcal{G}_j} = \arg\min_{^p\mathcal{G}_j\in \mathcal{P}_i}\sum_{k^\prime}\|\mathbf{e}_\text{proj}\left(\hat{\mathbf{x}}^j_i, \boldsymbol{\xi}_{k^\prime}\right)\|_{\boldsymbol{\Sigma}_{ik^\prime}}.$$

As our method in \autoref{sec.proj} still can not guarantee the map is projected perfectly, mis-projection of components are inevitable.
We further verify the likelihood of $\hat{\mathbf{x}}_i$ and then follow an ICP scheme to re-generate the association if the likelihood is low (line 10-20).
The procedure can be decomposed into the following two procedures:
\begin{enumerate}
	\item Given $\hat{\mathbf{x}}_i$, compute $\log(p(\hat{\mathbf{x}}_i | \mathcal{G}_h))$ for $\mathcal{G}_h \in \mathbf{n}(\hat{\mathcal{G}_j})$. $\mathbf{n}(\hat{\mathcal{G}_j})$ stands for the set of $\hat{\mathcal{G}_j}$'s neighbours, with the distance defined as $dist_{3}(\hat{\mathcal{G}_j}, \mathcal{G}_h)$.
	      Then the component with maximum likelihood is assigned to $\hat{\mathcal{G}_j}$.
	\item Recompute \autoref{eq.str_opt} to get $\hat{\mathbf{x}}_i$.
\end{enumerate}

We iterate until the likelihood of $\mathbf{x}_i$ given $\hat{\mathcal{G}_j}$ is the largest compared to all its neighbours.
In this way, the final association not only minimize the reprojection error, but also maximize the likelihood of the landmark.

\begin{algorithm}[t]
	\caption{Association and Structure Optimization}
	\label{alg.alg_association}
	\providecommand{\DontPrintSemicolon}{\dontprintsemicolon}
	\KwData{$\mathbf{u}_{ik}, {\boldsymbol{\xi}}_k, \mathbf{x}_i$.}

	$\mathcal{P}_i$ = \FuncSty{candidatesFromProjections($\mathbf{u}_{ik}$,$\mathcal{P}$)} 

	$\hat{^p\mathcal{G}_j} \leftarrow$\textit{null},  $\hat{e}_\text{proj} \leftarrow \infty$

	\For(){$^p\mathcal{G}_j$ in $\mathcal{P}_i$}
	{
		$b_\text{opt}$, $\hat{\mathbf{x}}_i$, $e_\text{proj}$ = \FuncSty{optStructure}($\mathbf{u}_{ik}$, ${\boldsymbol{\xi}}_k$, $\mathcal{G}_j$)

		\CommentSty{/* $b_\text{opt}$: flag for convergence. */}

		\If(){$e_\text{proj} <$ \FuncSty{min}$(th, \hat{e}_\text{proj})$}
		{
			$\hat{^p\mathcal{G}_j} \leftarrow$$^p\mathcal{G}_j$,  $\hat{e}_\text{proj} \leftarrow e_\text{proj}$

		}
	}
	\eIf(){$\hat{^p\mathcal{G}_j} \ne$null}{

		\SetKwRepeat{Repeat}{do}{while}%

		\Repeat(){$\log p(\hat{\mathbf{x}}_i | \hat{\mathcal{G}}_h) > \log p(\hat{\mathbf{x}}_i | \hat{\mathcal{G}}_j)$}{

			$^p\hat{\mathcal{G}}_j \leftarrow \hat{\mathcal{G}}_h$

			$b_\text{opt}$, $\hat{\mathbf{x}}_i$, $e_\text{proj}$ = \FuncSty{optStructure}($\mathbf{u}_{ik}$, ${\boldsymbol{\xi}}_k$, $\hat{\mathcal{G}}_j$)
			\If(){$b_\text{opt}$}{
				$\mathcal{M}_j$ = \FuncSty{findNeighbours}($\hat{^p\mathcal{G}_j}$)

				$\hat{\mathcal{G}}_h$ = $\arg \min_{\mathcal{G}_h \in \mathcal{M}_j} \log p(\hat{\mathbf{x}}_i | \mathcal{G}_h)$

			}
		}

	}
	(){
		\Return{false, $\mathbf{x}_i$, null}
	}

	\Return{true, $\hat{\mathbf{x}}_i$, $\hat{\mathcal{G}}_j$}
\end{algorithm}

\subsection{Joint Optimization}
With the map constraints, \autoref{eq.opt_total} is minimized to solve both keyframe poses and local structure.
Similar to \cite{mur2017orb},
the problem is solved by the Levenberg-Marquardt method, which gives a system as follows:
\begin{equation}
	\mathbf{H}=\mathbf{J}^{T} \mathbf{W} \mathbf{J}+ \epsilon \mathbf{I}, \quad  \quad \mathbf{b}=-\mathbf{J}^{T} \mathbf{W} \mathbf{r},
\end{equation}
where $\mathbf{J}$ and $\mathbf{r}$ are the stacked Jacobians and residuals, respectively. $\mathbf{W}$ is the weight matrix from stacking the inverse of the covariance for different residual terms as in \autoref{sec.residual}.
During the optimization, we further use $\mathcal{X}^2$-test with $95\%$ confidence to filter outliers and perform another round of optimizations with the outliers discarded.

\subsection{Other Implementation Details}
\subsubsection{Map Processing}
We train the GMM from the raw point cloud.
The number of total components varies according to different scenes.
When initializing the localization system, we load the offline constructed GMM map, decompose the covariance of all the components and check whether they are degenerated.
Neighbourhoods are also defined in this procedure.

\begin{table*}[t!]
	\caption{
		Left: Evaluation of the localization performance on the EuRoC Mav dataset.
		We report average Absolute Trajectory Error(ATE) (m) \cite{sturm2012benchmark} for 5 runs on EuRoC MAV dataset.
		The \textbf{best} and \underline{second-best} results are highlighted and the \sout{metric} striked out stands for a partial trajectory.
		For HFNet, $(\cdot)$ after ATE is the total number of failure frames. Right: \textbf{Settings} of the different methods in the experiment.
	}
	\label{tab.ate_euroc}
	\centering
	\begin{tabular}{ c c c c c c c }
		\toprule
		\multirow{2}{*}{\textbf{Seq.}} & \multirow{2}{*}{\textbf{Ours}} & \multirow{2}{*}{\textbf{DSL}} & \textbf{MSCKF}  & \multirow{2}{*}{\textbf{VINS-Mono}} & \multirow{2}{*}{\textbf{ORB-SLAM2}} & \multirow{2}{*}{\textbf{HFNet}} \\ \Tstrut{}
		                               &                                &                               & \textbf{w/ map}                                                                                                               \\ \midrule
		V101                           & \textbf{0.030}                 & 0.035                         & 0.056           & 0.044                               & \underline{0.033}                      & 0.062 (3)                       \\
		V102                           & \textbf{0.023}                 & \underline{0.034}                & 0.055           & 0.054                               & 0.047                               & -                               \\
		V103                           & \underline{0.047}                 & \textbf{0.045}                & 0.087           & 0.209                               & 0.199                               & 0.118 (79)                      \\
		V201                           & \textbf{0.018}                 & \underline{0.026}                & 0.069           & 0.062                               & 0.040                               & 0.083 (10)                      \\
		V202                           & \textbf{0.020}                 & \underline{0.023}                & 0.089           & 0.114                               & 0.065                               & -                               \\

		V203                           & \sout{0.056}              & \textbf{0.103}                & \underline{0.149}  & \underline{0.149}                      & \sout{0.242}                   & 0.117 (153)                     \\
		\bottomrule
	\end{tabular} \quad
	\begin{tabular}{ @{}ccc@{} }
		\toprule
		\textbf{Method} & \textbf{Input Sensors} & \textbf{Prior Map} \\ 
		\midrule
		ORB-SLAM2       & Stereo                 & (Not Use)          \\
		VINS-Mono       & Mono + IMU             & (Not Use)          \\
		MSCKF (w/ map)  & Stereo + IMU           & Point Cloud        \\
		DSL             & Mono                   & Surfel Map         \\
		HFNet           & Mono                   & SfM (sparse)       \\
		Ours            & Stereo                 & GMM                \\
		\bottomrule
	\end{tabular}
	\vspace{-1em}
\end{table*}

\subsubsection{Visual Tracking}
Here we follow ORB-SLAM2 \cite{mur2017orb}, an indirect visual SLAM method for camera tracking.
Briefly, the frontend extracts ORB features \cite{rublee2011orb} in the incoming frame and associates them with landmarks observed in the previous frame and map. Then, the current camera pose is recovered in a Perspective-n-Point (PnP) scheme.
After the initial tracking, the frontend decides whether to insert a keyframe into the backend mainly based on the current tracking quality.
The proposed method above happens right after a keyframe is inserted into the backend.

\subsubsection{Backend Management}
Our localization module maintains a local covisibility map, keeps merging similar landmarks and deleting redundant keyframes.
For the details of frontend tracking and backend management, we refer the readers to \cite{mur2017orb}.


\section{Experimental Results}
\label{sec:experiments}

We validate the proposed system on the public EuRoC MAV dataset \cite{burri25012016}. It provides sequences of stereo images and IMU data streams in three different indoor scenes, with extrinsic calibration, ground truth trajectories and dense reconstruction for two Vicon room configurations (denoted as V1, V2).
The main advantages of this dataset are two-fold:
first, six sequences including dense scene structures, which supports both cross-modality localization and reconstruction evaluation;
second, the aggressive motions and inconsistent illuminations bringing significant challenges for the visual state estimation.

We first evaluate the general localization performance against several state-of-the-art visual or visual-inertial state estimators.
Then, we dive into how introducing structure constraints can improve localization performance through the visual structure evaluation, which is followed by a parameter study.
Finally, the timing results are provided to prove the real-time performance.
All the experiments are performed using a desktop computer equipped with an Intel i7-8700K CPU and 16 GB RAM.

\subsection{General Localization Performance}

We compare our method, GMMLoc, with 5 state-of-the-art visual state estimators: our previous work DSL \cite{ye2020monocular}, MSCKF with pre-built map (w/ map) \cite{zuo2019visual}, VINS-Mono \cite{qin2018vins}, ORB-SLAM2 \cite{mur2015orb} and HFNet \cite{sarlin2019coarse}.
Among all the methods, GMMLoc, DSL, and MSCKF (w/ map) are similar in that they explicitly introduce the prior-map constraints into a visual state estimation, which we categorize as \textit{dense structure-based} localization methods.
On the contrary, HFNet is one the state-of-the-art \textit{sparse structure-based} localization methods, which follows a \textit{SfM-then-localization} pipeline.
We also compare the performance of VINS-Mono and ORB-SLAM2, the state-of-the-art VIO/VSLAM methods, following the evaluation protocol in \cite{zuo2019visual, ye2020monocular}.
We further list the settings of different methods in \autoref{tab.ate_euroc}.
Additionally, to show how the proposed method can improve the estimation accuracy of pure visual odometry, detecting the loop-closure from the image similarity is disabled in the comparison.
To evaluate HFNet, we first reconstruct the sparse feature map from SfM on one sequence and then perform localization on the other two under the same scene configuration.
We found that the images captured in two \textit{medium} sequences are of the best quality and thus provide a better SfM model for localization.

\begin{figure}[t!]
	\centering
	\includegraphics[width=0.48\textwidth]{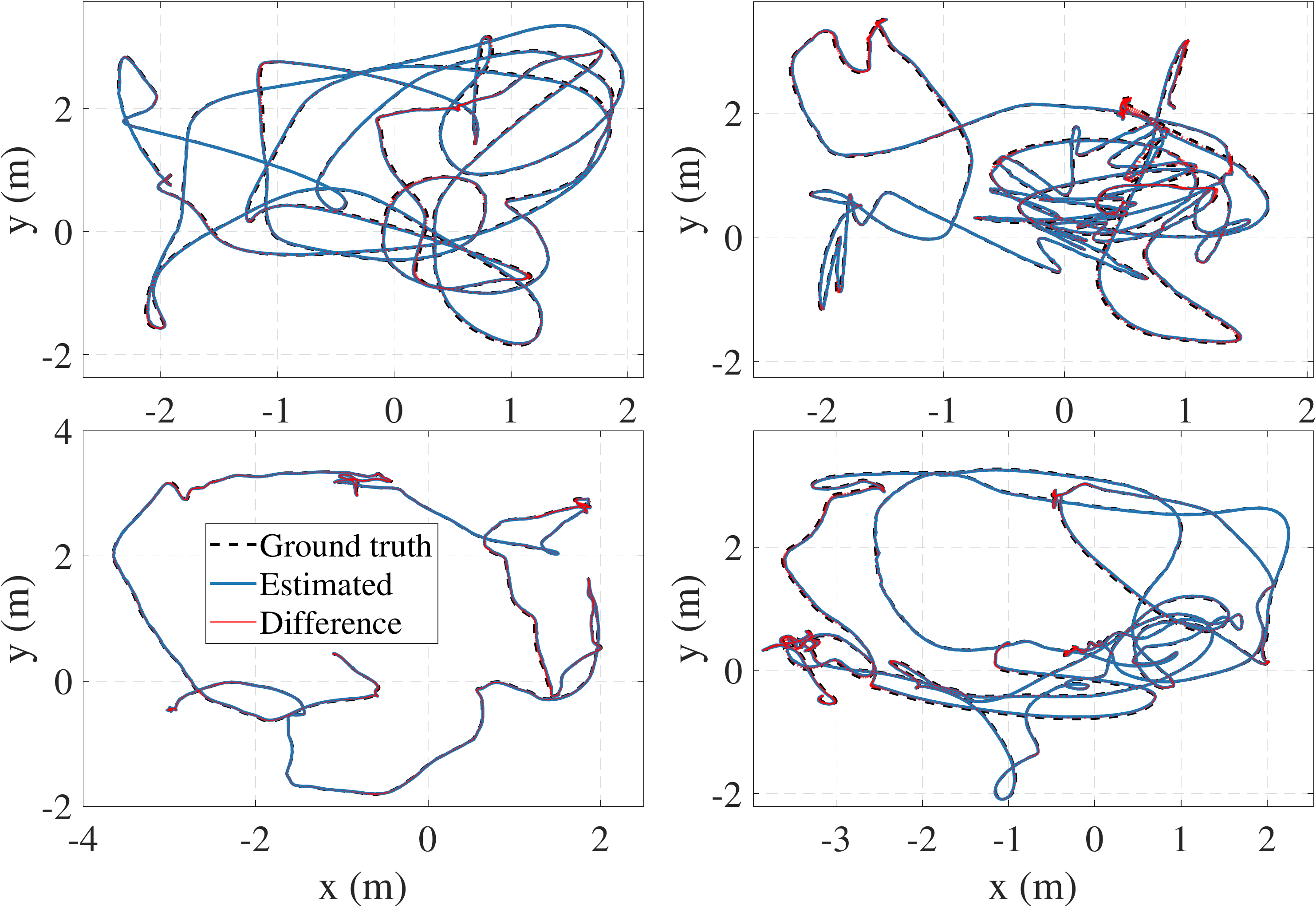}
	\caption{Qualitative comparison of the estimated trajectory and ground truth on sequences V102 (top left), V103 (top right), V201 (bottom left), V202 (bottom right).}
	\vspace{-1em}
	\label{fig.traj_qualitative}
\end{figure}

The localization results are presented in \autoref{tab.ate_euroc}.
Generally, our method achieves accurate estimation results compared to other methods.
A failure case occurs on sequence V203, where due to the lack of more than 300 frames of the left camera, the indirect frontend in our system, which is similar to our baseline method ORB-SLAM2, cannot manage to track consistently.
Therefore, only the accuracy of a partial trajectory is reported.
However, we still observe that our method corrects the tracking drift with structure constraints.

Compared to our previous work DSL, our system achieves a comparable localization accuracy.
It is also notable that performance degradation occurs mainly on the difficult sequences.
The reasons are two-fold:
first, both methods rely on the \textit{projection} procedure for the association, thus the tracking accuracy of the frontend has a significant effect on the association precision;
second, our method aims to make a trade-off between accuracy and efficiency,
as a consequence of which modelling the scene structure as a GMM does lose some of the structural information.
In addition, our method outperforms MSCKF (w/ map) in localization accuracy, while we do not densely reconstruct the scene structure from multiview stereo.
Compared to VINS-Mono or ORB-SLAM2, which uses visual-inertial or pure visual information, our method introduces structural constraints and generally improves the localization performance.
We also provide some qualitative results in \autoref{fig.traj_qualitative}. As shown in the figure, the localization with temporal visual constraints generally has no drift and even the maximum localization error is within an acceptable range (10-20 cm as visualized in the figure).

\subsection{Evaluations of the Local Reconstruction}

We evaluate the local structure reconstruction results of our method and ORB-SLAM2 using the ground truth 3-D model provided by the EuRoC Mav dataset.
The sparse feature maps generated by the state estimators are aligned and transformed under the map coordinate.
The error metric is defined as the RMSE of distances to the nearest neighborhood.
A similar evaluation process can be found in \cite{whelan2015elasticfusion}.
Five sequences on which both methods succeed are selected for the evaluation.
\begin{figure}[t!]
	\centering
	\includegraphics[width=0.35\textwidth]{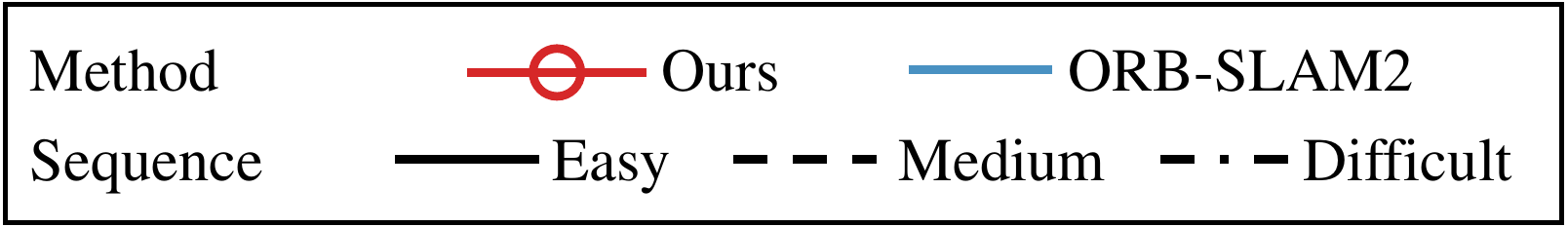}

	\vspace{1em}

	\advance\leftskip-1em
	\includegraphics[width=0.5\textwidth]{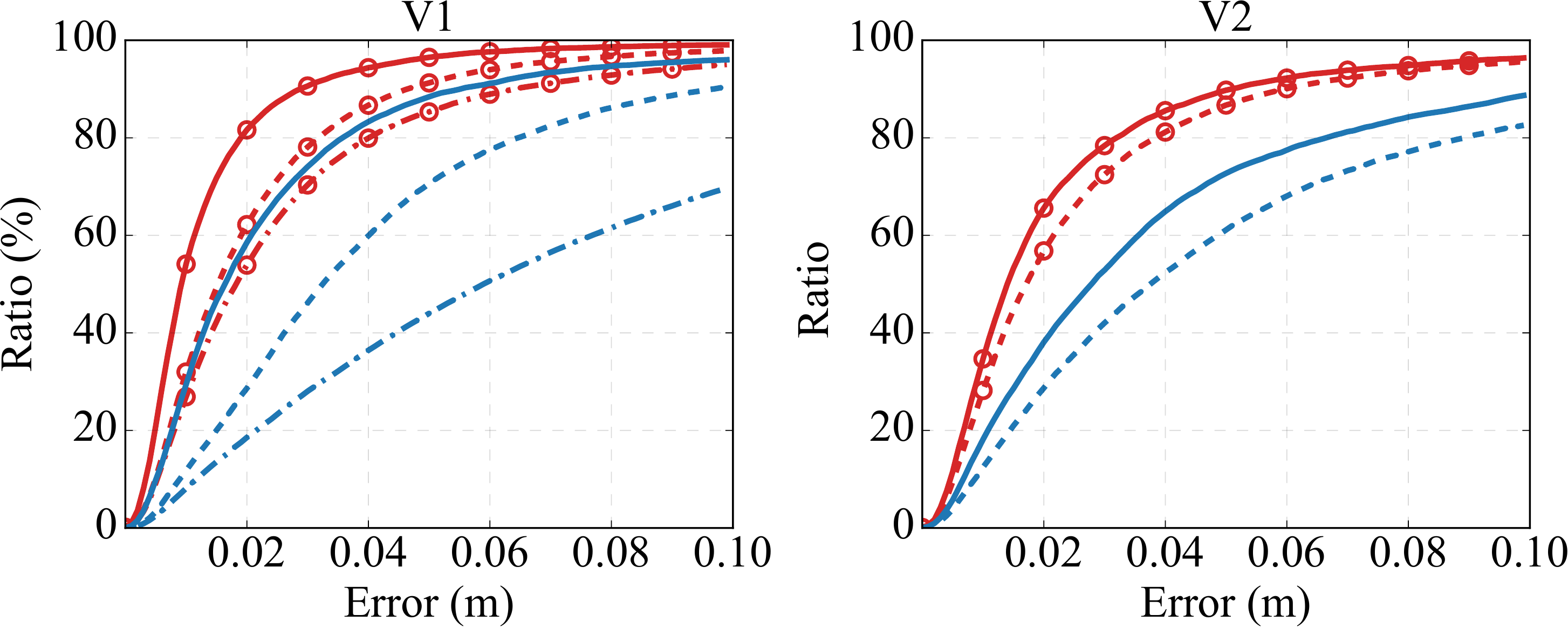}
	\caption{Evaluation of the visual structure accuracy estimated by Ours and ORB-SLAM2 on different sequences of V1 (left) and V2 (right).}
	\label{fig.eval_map}
\end{figure}

\begin{table}[t!]
	\caption{Average Root Mean Square Error (RMSE) (cm) of the local reconstruction, lower the better ($\downarrow$).}
	\label{tab.eval_map_metric}
	\centering
	\begin{tabular}{ @{}cccccc@{} }
		\toprule
		\textbf{Method} & \textbf{v101} & \textbf{v102} & \textbf{v103} & \textbf{v201} & \textbf{v202} \\ 
		\midrule
		Ours            & 0.098         & 0.184         & 2.711         & 0.408         & 0.934         \\
		ORB-SLAM2       & 0.338         & 0.706         & 6.300         & 1.620         & 1.880         \\
		\bottomrule
	\end{tabular}
	\vspace{-1em}
\end{table}

We present the ratio of inliers given an error threshold in \autoref{fig.eval_map} and report the metric results in \autoref{tab.eval_map_metric}. As shown in both \autoref{fig.eval_map} and \autoref{tab.eval_map_metric}, our method recovers a more accurate local structure, which in turn guarantees the accuracy of local trajectory estimation.
Noticeably, as the sequence becomes more challenging, the drawback of pure VO occurs. The estimation drift of VO is not negligible and it can not maintain a globally consistent visual structure.
In addition, even under V101 where both methods achieves similar localization performances, our method still outperforms the baseline in terms of structure accuracy.
This indicates that introducing the scene structure can also help our system filter the outliers out in two aspects: first, the poses are more accurate in our system, therefore outliers with a larger error in the visual factors, can be more easily detected;
second, the BA is constrained by the structure factors, thus the consistency of the visual structure is always maintained.

\begin{figure}[t!]
	\centering
	\includegraphics[width=0.48\textwidth]{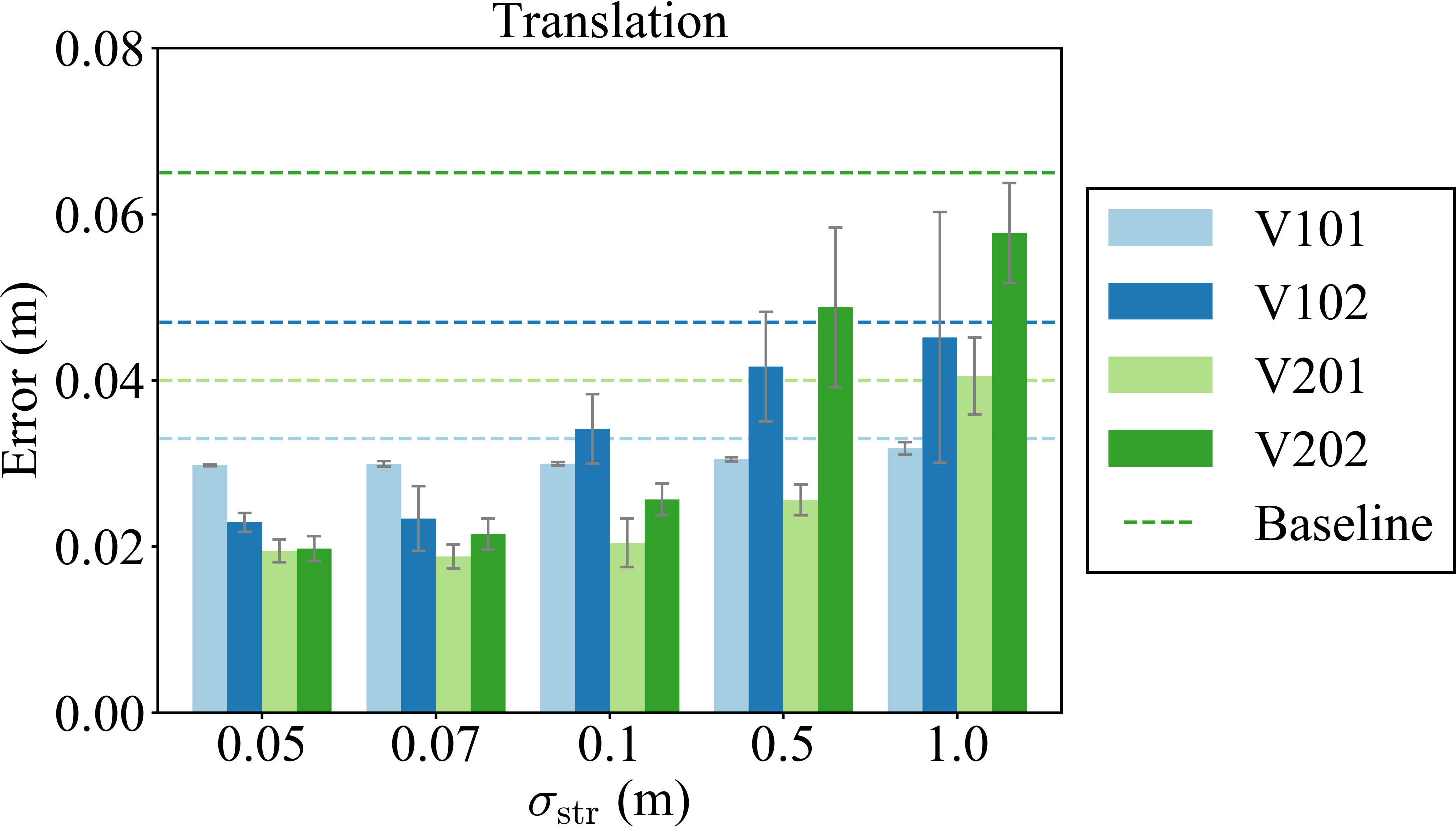}
	\caption{Parameter study on $\boldsymbol{\sigma}_\text{str}$. Localization error (rectangular bar) with variances (grey error bar) w.r.t $\sigma_\text{str}$ on 4 sequences are shown. The average ATE of ORB-SLAM2 is shown as dashed lines. Error on different sequences are distinguished by different colors.}
	\label{fig.param}
\end{figure}

\begin{table}[t!]
	\caption{Timings for the different modules of GMMLoc, are all tested in a single thread. The overheads compared to pure visual odometry are \textbf{highlighted}. Note that the Map Preprocessing is performed only at the start of the system.}
	\label{tab.timing}
	\centering
	\begin{tabular}{@{}llc@{}}
		\toprule
		\multicolumn{2}{c}{\textbf{Module}} & \textbf{Time (ms)}                                        \\ \midrule
		Initialization                      & \textbf{Map Preprocessing}      & \textbf{15.4 $\pm$ 0.0} \\ \midrule
		\multirow{5}{*}{Tracking}           & Feature Extraction              & 19.1 $\pm$ 3.4          \\
		                                    & Camera Tracking                 & 15.1 $\pm$ 1.6          \\
		                                    & KeyFrame Creation               & 0.1 $\pm$ 0.3           \\
		                                    & Total                           & 34.5 $\pm$ 4.5          \\ \midrule
		\multirow{6}{*}{Localization}
		                                    & \textbf{Map Projection}         & \textbf{3.5 $\pm$  3.2} \\
		                                    & \textbf{Initial Association   } & \textbf{0.5 $\pm$  0.3} \\
		                                    & \textbf{Structure Optimization} & \textbf{8.2 $\pm$  2.3} \\
		                                    & Local BA                        & 350.5 $\pm$  201.5      \\
		                                    & Map Management                  & 52.3 $\pm$  12.6        \\
		                                    & Total                           & 409.7 $\pm$ 205.5       \\ \bottomrule
	\end{tabular}
	\vspace{-1em}
\end{table}

\subsection{Effectiveness of the Structure Factor}
\label{sec.exp_param}
In this section, we further study how the parameters coupling the structure constraints with visual constraints influence the localization performance.
For different values of ${\sigma_\text{str}}$, we perform 5 Monte Carlo runs on each sequence,
and
report the average ATE with variances in \autoref{fig.param}.
By increasing ${\sigma}_\text{str}$, the average localization error approaches to that of the baseline method.
This provides an alternative view of
the improvement in the localization accuracy compared to pure visual odometry by introducing structure constraints.
Especially when ${\sigma}_\text{str} = 1$ m, there is only a trivial improvement to the localization accuracy (even very close to that of the baseline on V201).
Note that if we simply discard the structure constraints, the performance should be the same with the baseline.
As ${\sigma}_\text{str}$ also represents how the optimization weighs structure constraints, we observe that generally, with a low ${\sigma}_\text{str}$ value (in our experiment, 0.05-0.1 m), the estimation can be more consistent over different runs (shown as low variances in \autoref{fig.param}).

\subsection{Runtime Analysis}

To demonstrate the real-time capability of the proposed method, we report the runtime analysis in \autoref{tab.timing}.
As mentioned in \autoref{sec:intro}, the projection and association can be rather efficient and the only trivial overhead is introduced to the vision-only backend.
Based on the evaluation, it only costs around 10ms, which takes up to around 1/40 in the backend optimization.
In addition, as such an overhead only occurs in the backend, the time cost can be even less than 1ms if averaged by the frame rate.
This gives solid support for the previous claim that our system is more efficient and has the potential to be applied to embedded platforms.

As reported in MSCKF (w/ map) \cite{zuo2019visual} and DSL \cite{ye2020monocular},
MSCKF (w/ map) performs dense reconstruction and NDT-based local registration to localize the camera, which achieves a frequency of around 1.25Hz,
while DSL utilizes a modern GPU to render the scene structure for the data association.
On the contrary, the proposed method projects the global map elements and associates them with local observations in a flash, using only a CPU without multi-threading.
Yet, our method still exhibits the capability of estimating an accurate trajectory,
indicating that it makes a good trade-off between accuracy and efficiency.




\section{Conclusions and Future Work}
In this letter, we presented a structure-consistent visual localization method using the GMM as a map representation.
Given a camera pose tracked by the indirect front-end, the GMM map is projected back via a non-linear Gaussian transform, and several criteria are applied for a photorealistic projection.
Association is performed in three hierarchical steps, searching candidates, finding the component to minimize the reprojection error, and further verify the association with likelihood.
In the meantime, the landmark position from triangulation is refined with structural constraints.
Finally, the back-end jointly optimizes the visual structure, and the keyframe poses.
The experimental results demonstrated the effectiveness of the proposed method.
As our method balances accuracy and efficiency well, we believe it has the potential to be applied to onboard platforms in the future.

As the next step, we plan to investigate how to initialize the depth of keypoints from a GMM projection. 
Additionally, we believe it is also worthy of studying how to introduce some high-level information like semantics to boost the system.
Last but not least, introducing IMU factors for a smoother and more robust pose estimation is also promising for increasing the general localization performance. 


\balance
\bibliography{main}

\begin{thebibliography}{10}
\providecommand{\url}[1]{#1}
\csname url@samestyle\endcsname
\providecommand{\newblock}{\relax}
\providecommand{\bibinfo}[2]{#2}
\providecommand{\BIBentrySTDinterwordspacing}{\spaceskip=0pt\relax}
\providecommand{\BIBentryALTinterwordstretchfactor}{4}
\providecommand{\BIBentryALTinterwordspacing}{\spaceskip=\fontdimen2\font plus
\BIBentryALTinterwordstretchfactor\fontdimen3\font minus
  \fontdimen4\font\relax}
\providecommand{\BIBforeignlanguage}[2]{{%
\expandafter\ifx\csname l@#1\endcsname\relax
\typeout{** WARNING: IEEEtran.bst: No hyphenation pattern has been}%
\typeout{** loaded for the language `#1'. Using the pattern for}%
\typeout{** the default language instead.}%
\else
\language=\csname l@#1\endcsname
\fi
#2}}
\providecommand{\BIBdecl}{\relax}
\BIBdecl

\bibitem{liu2014topo}
M.~{Liu} and R.~{Siegwart}, ``Topological mapping and scene recognition with
  lightweight color descriptors for an omnidirectional camera,'' \emph{IEEE
  Transactions on Robotics}, vol.~30, no.~2, pp. 310--324, 2014.

\bibitem{desouza2002vision}
G.~N. DeSouza and A.~C. Kak, ``Vision for mobile robot navigation: A survey,''
  \emph{IEEE transactions on pattern analysis and machine intelligence},
  vol.~24, no.~2, pp. 237--267, 2002.

\bibitem{mur2017orb}
R.~Mur-Artal and J.~D. Tard{\'o}s, ``{ORB-SLAM2}: An open-source slam system
  for monocular, stereo, and rgb-d cameras,'' \emph{IEEE Transactions on
  Robotics}, vol.~33, no.~5, pp. 1255--1262, 2017.

\bibitem{pascoe2015direct}
G.~Pascoe, W.~Maddern, and P.~Newman, ``Direct visual localisation and
  calibration for road vehicles in changing city environments,'' in \emph{the
  IEEE International Conference on Computer Vision Workshops}, 2015, pp. 9--16.

\bibitem{caselitz2016mloc}
T.~Caselitz, B.~Steder, M.~Ruhnke, and W.~Burgard, ``Monocular camera
  localization in 3-d lidar maps,'' in \emph{2016 IEEE/RSJ International
  Conference on Intelligent Robots and Systems (IROS)}.\hskip 1em plus 0.5em
  minus 0.4em\relax IEEE, 2016, pp. 1926--1931.

\bibitem{kim2018stereo}
Y.~Kim, J.~Jeong, and A.~Kim, ``Stereo camera localization in 3-d lidar maps,''
  in \emph{2018 IEEE/RSJ International Conference on Intelligent Robots and
  Systems (IROS)}.\hskip 1em plus 0.5em minus 0.4em\relax IEEE, 2018, pp. 1--9.

\bibitem{ding2018laser}
X.~Ding, Y.~Wang, D.~Li, L.~Tang, H.~Yin, and R.~Xiong, ``Laser map aided
  visual inertial localization in changing environment,'' in \emph{2018
  IEEE/RSJ International Conference on Intelligent Robots and Systems
  (IROS)}.\hskip 1em plus 0.5em minus 0.4em\relax IEEE, 2018, pp. 4794--4801.

\bibitem{huang2019metric}
H.~{Huang}, Y.~{Sun}, H.~{Ye}, and M.~{Liu}, ``Metric monocular localization
  using signed distance fields,'' in \emph{2019 IEEE/RSJ International
  Conference on Intelligent Robots and Systems (IROS)}, 2019, pp. 1195--1201.

\bibitem{zuo2019visual}
X.~Zuo, P.~Geneva, Y.~Yang, W.~Ye, Y.~Liu, and G.~Huang, ``Visual-inertial
  localization with prior lidar map constraints,'' \emph{IEEE Robotics and
  Automation Letters}, vol.~4, no.~4, pp. 3394--3401, 2019.

\bibitem{ye2020monocular}
H.~Ye, H.~Huang, and M.~Liu, ``Monocular direct sparse localization in a prior
  3-d surfel map,'' in \emph{2020 IEEE International Conference on Robotics and
  Automation (ICRA)}.\hskip 1em plus 0.5em minus 0.4em\relax IEEE, 2020.

\bibitem{gold1998new}
S.~Gold, A.~Rangarajan, C.-P. Lu, S.~Pappu, and E.~Mjolsness, ``New algorithms
  for 2-d and 3-d point matching: Pose estimation and correspondence,''
  \emph{Pattern recognition}, vol.~31, no.~8, pp. 1019--1031, 1998.

\bibitem{reynolds2009gaussian}
D.~A. Reynolds, ``Gaussian mixture models.'' \emph{Encyclopedia of biometrics},
  vol. 741, 2009.

\bibitem{mur2015orb}
R.~Mur-Artal, J.~M.~M. Montiel, and J.~D. Tardos, ``{ORB-SLAM}: a versatile and
  accurate monocular slam system,'' \emph{IEEE Transactions on Robotics},
  vol.~31, no.~5, pp. 1147--1163, 2015.

\bibitem{besl1992method}
P.~J. Besl and N.~D. McKay, ``Method for registration of 3-d shapes,'' in
  \emph{Sensor fusion IV: control paradigms and data structures}, vol.
  1611.\hskip 1em plus 0.5em minus 0.4em\relax International Society for Optics
  and Photonics, 1992, pp. 586--606.

\bibitem{rusinkiewicz2001efficient}
S.~Rusinkiewicz and M.~Levoy, ``Efficient variants of the icp algorithm,'' in
  \emph{Proceedings Third International Conference on 3-D Digital Imaging and
  Modeling}.\hskip 1em plus 0.5em minus 0.4em\relax IEEE, 2001, pp. 145--152.

\bibitem{myronenko2010point}
A.~Myronenko and X.~Song, ``Point set registration: Coherent point drift,''
  \emph{IEEE transactions on pattern analysis and machine intelligence},
  vol.~32, no.~12, pp. 2262--2275, 2010.

\bibitem{gao2019filterreg}
W.~Gao and R.~Tedrake, ``Filterreg: Robust and efficient probabilistic
  point-set registration using gaussian filter and twist parameterization,'' in
  \emph{Proceedings of the IEEE Conference on Computer Vision and Pattern
  Recognition}, 2019, pp. 11\,095--11\,104.

\bibitem{eckart2018eccv}
B.~Eckart, K.~Kim, and J.~Kautz, ``Hgmr: Hierarchical gaussian mixtures for
  adaptive 3-d registration,'' in \emph{The European Conference on Computer
  Vision (ECCV)}, September 2018.

\bibitem{dhawale2018fast}
A.~Dhawale, K.~Shaurya~Shankar, and N.~Michael, ``Fast monte-carlo localization
  on aerial vehicles using approximate continuous belief representations,'' in
  \emph{Proceedings of the IEEE Conference on Computer Vision and Pattern
  Recognition}, 2018, pp. 5851--5859.

\bibitem{o2018variable}
C.~O’Meadhra, W.~Tabib, and N.~Michael, ``Variable resolution occupancy
  mapping using gaussian mixture models,'' \emph{IEEE Robotics and Automation
  Letters}, vol.~4, no.~2, pp. 2015--2022, 2018.

\bibitem{corah2019communication}
M.~Corah, C.~O’Meadhra, K.~Goel, and N.~Michael, ``Communication-efficient
  planning and mapping for multi-robot exploration in large environments,''
  \emph{IEEE Robotics and Automation Letters}, vol.~4, no.~2, pp. 1715--1721,
  2019.

\bibitem{barfoot2019state}
T.~D. Barfoot, ``State estimation for robotics,'' 2019.

\bibitem{sturm2012benchmark}
J.~Sturm, N.~Engelhard, F.~Endres, W.~Burgard, and D.~Cremers, ``A benchmark
  for the evaluation of rgb-d slam systems,'' in \emph{2012 IEEE/RSJ
  International Conference on Intelligent Robots and Systems}.\hskip 1em plus
  0.5em minus 0.4em\relax IEEE, 2012, pp. 573--580.

\bibitem{rublee2011orb}
E.~Rublee, V.~Rabaud, K.~Konolige, and G.~Bradski, ``Orb: An efficient
  alternative to sift or surf,'' in \emph{2011 International conference on
  computer vision}.\hskip 1em plus 0.5em minus 0.4em\relax Ieee, 2011, pp.
  2564--2571.

\bibitem{burri25012016}
\BIBentryALTinterwordspacing
M.~Burri, J.~Nikolic, P.~Gohl, T.~Schneider, J.~Rehder, S.~Omari, M.~W.
  Achtelik, and R.~Siegwart, ``The euroc micro aerial vehicle datasets,''
  \emph{The International Journal of Robotics Research}, 2016. [Online].
  Available:
  \url{http://ijr.sagepub.com/content/early/2016/01/21/0278364915620033.abstract}
\BIBentrySTDinterwordspacing

\bibitem{qin2018vins}
T.~Qin, P.~Li, and S.~Shen, ``{VINS-MONO}: A robust and versatile monocular
  visual-inertial state estimator,'' \emph{IEEE Transactions on Robotics},
  vol.~34, no.~4, pp. 1004--1020, 2018.

\bibitem{sarlin2019coarse}
P.-E. Sarlin, C.~Cadena, R.~Siegwart, and M.~Dymczyk, ``From coarse to fine:
  Robust hierarchical localization at large scale,'' in \emph{Proceedings of
  the IEEE Conference on Computer Vision and Pattern Recognition}, 2019, pp.
  12\,716--12\,725.

\bibitem{whelan2015elasticfusion}
T.~Whelan, S.~Leutenegger, R.~Salas-Moreno, B.~Glocker, and A.~Davison,
  ``Elasticfusion: Dense slam without a pose graph.''\hskip 1em plus 0.5em
  minus 0.4em\relax Robotics: Science and Systems, 2015.

\end{thebibliography}
\bibliographystyle{IEEEtran}

\addtolength{\textheight}{-12cm}   

\end{document}